\documentclass[10pt,twocolumn,letterpaper]{article}

\usepackage{cvpr}      
\usepackage{multirow}
\usepackage{cuted}
\usepackage{float}
\usepackage[dvipsnames]{xcolor}

\usepackage{amsmath,amsfonts,bm}
\usepackage{lipsum}
\usepackage{mathtools}

\newcommand{\bx}{\mathbf{x}}
\newcommand{\bz}{\mathbf{z}}

\newcommand{\bI}{\mathbf{I}}

\newcommand{\bv}{\mathbf{v}}

\newcommand{\defeq}{\coloneqq}
\newcommand{\beps}{\bm{\epsilon}}

\newlength\paramargin
\newlength\figmargin

\newlength\secmargin
\newlength\figcapmargin
\newlength\tabcapmargin

\newcommand{\topic}[1]
{
\vspace{1.5mm}\noindent\textbf{#1}
}

\newcommand{\tabref}[1]{Table~\ref{#1}}

\long\def\ignorethis#1{}

\newbox\jsavebox%

\makeatletter
\newcommand{\providelength}[1]{%
  \@ifundefined{\expandafter\@gobble\string#1}
   {
    \typeout{\string\providelength: making new length \string#1}%
    \newlength{#1}%
   }
   {
    \sdaau@checkforlength{#1}%
   }%
}

\def\eps{\bm{\epsilon}}

\definecolor{cvprblue}{rgb}{0.21,0.49,0.74}
\usepackage[pagebackref,breaklinks,colorlinks,citecolor=cvprblue]{hyperref}

\def\paperID{97} 
\def\confName{3DV\xspace}
\def\confYear{2025\xspace}
\def\sysname{GECO}

\title{\sysname{}: \underline{G}enerative Image-to-3D within a S\underline{ECO}nd}

\author{Chen Wang$^{1}$ \;\;
Jiatao Gu$^{2}$ \;\;
Xiaoxiao Long$^{3}$  \;\; Yuan Liu$^{3}$  \;\; Lingjie Liu$^{1}$\\
$^{1}$University of Pennsylvania\quad$^{2}$Apple\quad$^{3}$The University of Hong Kong}

\begin{document}
\maketitle


\begin{strip}
    \centering
    \vspace{-5em}
    \centering
    \includegraphics[width=\textwidth]{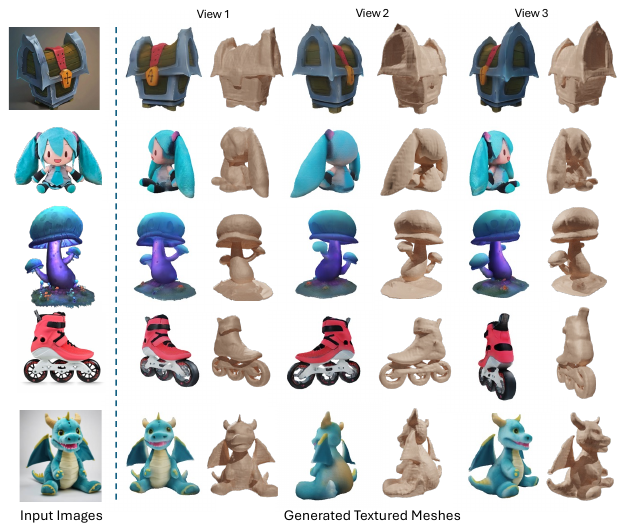}
    \vspace{-2em}
    \captionof{figure}{We propose \sysname{}, a framework for feed-forward image-to-3D generation that produces texture meshes in \textbf{0.64s} on a single L40 GPU. Here we show both the texture and geometry renderings of the generated meshes.}
    \label{fig:teaser}
\end{strip}

\begin{abstract}
Recent years have seen significant advancements in 3D generation. While methods like score distillation achieve impressive results, they often require extensive per-scene optimization, which limits their time efficiency. On the other hand, reconstruction-based approaches are more efficient but tend to compromise quality due to their limited ability to handle uncertainty. We introduce \sysname{}, a novel method for high-quality 3D generative modeling that operates within a second. Our approach addresses the prevalent issues of uncertainty and inefficiency in existing methods through a two-stage approach. In the first stage, we train a single-step multi-view generative model with score distillation. Then, a second-stage distillation is applied to address the challenge of view inconsistency in the multi-view generation. This two-stage process ensures a balanced approach to 3D generation, optimizing both quality and efficiency. Our comprehensive experiments demonstrate that \sysname{} achieves high-quality image-to-3D mesh generation with an unprecedented level of efficiency. We will make the code and model publicly available.
\end{abstract}

\section{Introduction}
\label{sec:intro}

3D digital assets encapsulate the geometry and appearance of objects from the real world. 
The role of 3D assets is pivotal across a wide range of applications, including movies, digital games, virtual reality, and robotics. Despite their importance, generating 3D assets is often labor-intensive and typically restricted to skilled professionals. 
Automatic and efficient techniques for generating high-fidelity 3D models will greatly simplify the workload and open up the creation process to beginners.
Therefore, in this paper, we study the problem of efficiently producing high-quality 3D assets using a single input image, aiming for fast and faithful reproduction of the original object in the image.



Dreamfusion~\cite{poole2022dreamfusion} and the follow-up works~\cite{lin2023magic3d, wang2023prolificdreamer, chen2023fantasia3d, qian2023magic123, liu2023zero, tang2023dreamgaussian} propose to distill 3D neural representations~\cite{mildenhall2021nerf, muller2022instant, kerbl20233dgs} from pretrained large-scale 2D diffusion models~\cite{rombach2022high,saharia2022photorealistic} with score distillation techniques. These methods generate high-quality 3D assets with text or image input, however, facing the major drawback that they require 30 minutes of per-scene optimization for only one object, which raises practical concerns in real-time applications.
On the other hand, to accelerate 3D generation, reconstruction-based models (e.g., PixelNeRF~\cite{yu2021pixelnerf}, LRM~\cite{hong2023lrm}, TripoSR~\cite{tochilkin2024triposr}) train a deterministic feed-forward model for predicting 3D representations given a single input image. By leveraging large-scale 3D datasets, such models exhibit impressive generalization ability over unseen objects, and only require less than a second to obtain the 3D.
However, the uncertainty issue of the single image to 3D prediction is fundamentally unsolvable for deterministic methods: unseen regions of a 3D object cannot be fully recovered from the single image input, causing blurriness and incorrect geometries.

Tackling the uncertainty issues, various methods have been proposed for incorporating generative models such as diffusion models for text-to-3D generation tasks~\cite{gu2023nerfdiff,li2023instant3d,tang2024lgm, wang2024crm}.
For instance, InstantMesh~\cite{xu2024instantmesh} employs a multi-view diffusion model~\cite{shi2023mvdream,shi2023zero123++} to first synthesize multi-view consistent images given a single image, and predict the final 3D representations based on the predicted images. In such case, the uncertainty problem can be addressed by the first-stage diffusion model, and as a result, the predicted 3D is generally better than pure reconstruction-based models. However, the multi-view diffusion stage still needs multiple network inferences and takes over 7 seconds due to the iterative sampling nature of diffusion models, and thus bottlenecks time-sensitive applications.

\begin{figure}[tp]
    \centering
    \includegraphics[width=1.0\linewidth]{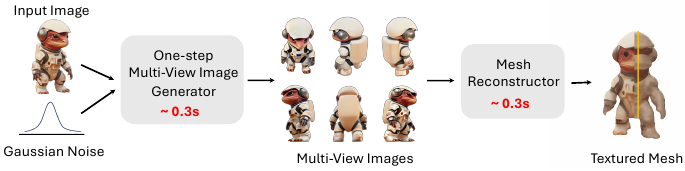}
    \caption{Overall pipeline of our feedforward 3D generator, which achieves image-to-3D mesh generation within one second given a conditional image and noise.}
    \label{fig:pipeline}
    \vspace{-20pt}
\end{figure}

To address these issues, we present \sysname{}, a generative approach that can generate high-fidelity 3D objects within one second. Specifically, we learn a feed-forward generator similar to reconstruction-based models~\cite{hong2023lrm,zou2023triplane} while taking additional noise as inputs for handling uncertainties. Training such a model from scratch is a non-trivial task due to mode collapse. Instead, we parameterize our model using multi-view images as an intermediate representation (\cref{fig:pipeline}), and propose a novel two-stage distillation approach for training (\cref{fig:learning_pipeline}).
For the first stage, we follow variational score distillation (VSD~\cite{wang2023prolificdreamer}) and learn a single-step multi-view generator directly from a pre-trained multi-view diffusion model~\cite{shi2023zero123++}.
Since the outputs of the single-step multi-view generator are not perfectly multi-view consistent, the reconstructed 3D model from the multi-view outputs tends to have incorrect geometry. 
To tackle this problem, we propose a second-stage training by fine-tuning a pretrained reconstruction-based method~\cite{xu2024instantmesh} with outputs of the single-step multi-view diffusion model from the first stage. We generate pseudo ground truth images using the multi-step diffusion model and the pretrained reconstruction model, which is more consistent, to train our second-stage model with reconstruction losses. Notably, this training strategy enables using images from arbitrary viewpoints to supervise high-quality reconstruction.

We conduct extensive quantitative and qualitative comparisons on GSO~\cite{downs2022google} dataset. We also test \sysname{} on more challenging in-the-wild input images (some of them are shown in \cref{fig:teaser}). The results show that our method can well resolve the uncertainty of image-to-3D generation, while being highly efficient in rendering and mesh extraction. Compared to previous feed-forward baselines, our method synthesizes high-quality texture and geometry even for the back view of the input object.

Our contributions can be summarized as the following:
\begin{itemize} 
    \item We design a novel feed-forward model for single-image-to-3D generation that, for the first time, handles the uncertainty issue while achieving high efficiency.
    \item We propose a two-stage distillation method that efficiently distills a pre-trained multi-view diffusion model and a reconstruction-based model into a feedforward image-to-3D generation model.
    \item Extensive experiments demonstrate that \sysname{} achieves high-quality 3D generation within one second, outperforming reconstruction-based methods in terms of quality and existing diffusion-based methods in generation speed.
\end{itemize}
\vspace{-10pt}
\section{Related Work}
\noindent{\bf{Acceleration of Diffusion Models}}
Diffusion Models~\cite{ho2020denoising, song2020score, song2020denoising, dhariwal2021diffusion}, also known as score-based generative models, achieved tremendous success for various generative tasks, including text~\cite{gong2022diffuseq}, image~\cite{rombach2022high, saharia2022photorealistic}, video~\cite{ho2022video, ho2022imagen} and 3D~\cite{gu2023nerfdiff, liu2023zero}. 
The continuous form of diffusion models are SDEs that transform between data distribution and a prior distribution~\cite{song2020score}. The SDEs also have corresponding probability flow ODE with the same marginal distribution~\cite{song2020score, song2020denoising}. One of the major drawbacks of diffusion models is that they require hundreds of denoising steps to generate the final output. Researchers have proposed efficient diffusion samplers~\cite{lu2022dpm, karras2022elucidating, zheng2024dpm, bao2022analytic} to reduce the sampling steps of pretrained diffusion models to less than 50. Another line of work formulates the acceleration problem under the framework of knowledge distillation~\cite{hinton2015distilling}, where a fast student model is distilled from the teacher model. The pioneering work of Salimans and Ho~\cite{salimans2022progressive} progressively reduces the number of steps for StableDiffusion by training multiple student models. Consistency models~\cite{song2023consistency, luo2023latent} and BOOT~\cite{gu2023boot} learn a one-step generator that matches the output of the teacher model along the ODE trajectory at each timestep by bootstrapping in a forward or backward manner respectively. Recently, ADD~\cite{sauer2023adversarial} and DMD~\cite{yin2023one} introduced score distillation~\cite{poole2022dreamfusion, wang2023prolificdreamer} for diffusion distillation.



\topic{3D Generation with Diffusion Models} 
Researchers have explored directly training diffusion models on 3D representations, \eg point clouds, triplanes, neural fields~\cite{luo2021diffusion, shue20233d, muller2023diffrf, chou2023diffusion, jun2023shap, wu2023hyperdreamer}. However, they require exhaustive 3D data and computation resources and are also limited to category-level shape generation with simple textures. Other works proposed to learn 3D models from 2D pretrained diffusion models with score distillation~\cite{poole2022dreamfusion, wang2023score, wang2023prolificdreamer} by matching the distribution of 3D renderings with that of 2D images.
Follow-up works further improves the quality by using high-resolution guidance~\cite{lin2023magic3d, chen2023fantasia3d, qian2023magic123,sun2023dreamcraft3d}, disentangling geometry and apperance~\cite{chen2023fantasia3d, qian2023magic123}, and introducing advanced diffusion guidance~\cite{metzer2023latent, zhu2023hifa, sun2023dreamcraft3d,liang2023luciddreamer,kwak2023vivid}.
Currently, these methods achieve high-fidelity 3D generation with detailed texture. Besides, the same objective is also widely utilized in scene-level generation~\cite{hollein2023text2room, po2023compositional}, 3D editing~\cite{zhuang2023dreameditor, li2023focaldreamer}, texturing~\cite{yeh2024texturedreamer, metzer2023latent} and articulated object generation~\cite{cao2023dreamavatar, jakab2023farm3d, liao2023tada}.

As an intermediate 3D representation, the generation of multi-view images using diffusion models has been explored. The advantage of multi-view images is that they are batched 2D projections and can be directly processed by existing image diffusion models with minor changes. Existing works~\cite{shi2023mvdream, liu2023syncdreamer, long2023wonder3d, shi2023zero123++,sargent2023zeronvs,lu2023direct2,shi2023toss,xie2023carve3d,tang2024mvdiffusion++} fine-tuned from pretrained StableDiffusion variants to generate view consistent multi-view images, which is then fused or reconstructed to 3D representations. However, they still require several seconds to perform diffusion sampling and our work addresses this by learning to generate multi-view images in one step.

\topic{Efficient 3D Generation} Methods based on score distillation often require several minutes of optimization to obtain one 3D model even with efficient 3D Gaussians~\cite{tang2023dreamgaussian, chen2023text}. Some works~\cite{lorraine2023att3d, qian2024atom} use score distillation to train a hypernetwork of neural fields, enabling 3D generation from direct inference but having limited generalization ability. 
Recently, LRM~\cite{hong2023lrm} and TripoSR~\cite{tochilkin2024triposr} trains a reconstruction model on large-scale datasets~\cite{deitke2023objaverse} and enable image-to-3D in seconds. TriplaneGaussian~\cite{zou2023triplane} uses 3D Gaussians to assist the generation of triplanes for LRM.
However, the major problem of reconstruction-based methods is that they do not consider the uncertain nature of 3D generation, so the back views of the generated objects are often blurry. Based on LRM, Instant3D~\cite{li2023instant3d} and InstantMesh~\cite{xu2024instantmesh} samples multi-view images with 2D diffusion for LRM reconstruction, and DMV3D~\cite{xu2023dmv3d} directly trains a 3D diffusion with LRM. All three works improve quality but sacrifice efficiency. 
\section{Preliminaries}
\begin{figure*}[htbp]
    \centering
    \includegraphics[width=1.0\linewidth]{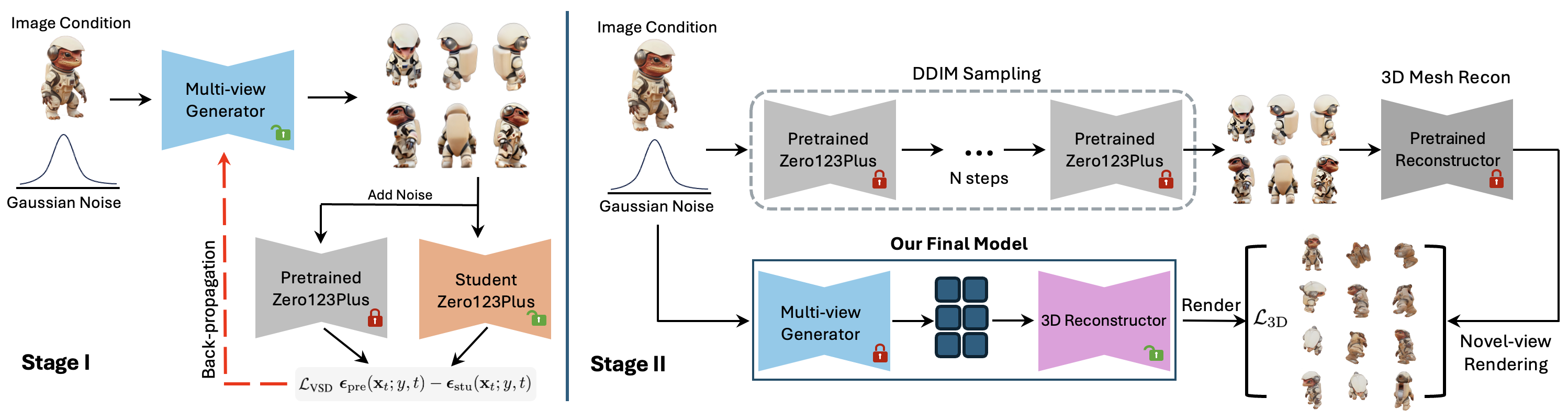}
    \vspace{-20pt}
    \caption{The two-stage learning pipeline for \sysname{}. Stage I: the multi-view generator is optimized with VSD~\cite{wang2023prolificdreamer} objective with a pre-trained multi-view diffusion model~\cite{shi2023zero123++}; Stage II: the full model is optimized by predicting the rendering from the pre-trained reconstruction model~\cite{tang2024lgm} under the same image and noise condition.}
    \vspace{-15pt}
    \label{fig:learning_pipeline}
\end{figure*}

\subsection{Multi-view Diffusion Models}
\label{subsec:pre}
\topic{Diffusion Models}~\cite{ho2020denoising, song2020denoising} learn the data distribution by estimating the noised data distribution (or score) along a Markov Chain. Diffusion models consist of a forward process that gradually removes information from data by adding Gaussian noises and a reverse process that generates data starting from random noise. Given $\bx_0 \sim q(\bx_0)$, the forward process $q$ is a Markov chain that adds gaussian noise to $\bx_0$ and generates latent $\bx_1, ..., \bx_{T}$ of the same dimension with $q(\bx_t | \bx) = \mathcal{N}(\alpha_t\bx, \sigma_t^2\mathbf{I})$. Ideally, the final latent $\bx_T$ will follow a standard Gaussian distribution: $p(\bx_T)=\mathcal{N}(\bx_T; \mathbf{0}, \mathbf{I})$. The reverse process starts denoising from $\bx_T$ by learning the Gaussian transitions from $\bx_t$ to $\bx_{t-1}$ that is defined as $p_\theta(\bx_{0:T}) \defeq p(\bx_T)\prod_{t=1}^T p_\theta(\bx_{t-1}|\bx_t)$. Further, $p_\theta(\bx_{t-1}|\bx_t) = \mathcal{N}(\bx_{t-1}, \mu_{\theta}(\bx_t, t), \sigma_t^2\bI)$ and $\mu_{\theta}$ is the learnable component. The sampling of diffusion models often takes more than 50 steps to obtain high-quality results.

\topic{Multi-view Diffusion Models} learn the joint probability distribution of multi-view images~\cite{liu2023syncdreamer, long2023wonder3d, shi2023zero123++}. This kind of model treats multi-view renderings of an object on a fixed set of viewpoints as the data point. The learning process of multi-view diffusion models is similar to standard image diffusion models except that noises are added and denoised simultaneously to those images. They also need special design to maintain the consistency of different viewpoints. However, the problem with these models is that multiple inference steps are required, and the results are not view-consistent enough. Furthermore, post-processing is also needed to reconstruct 3D geometry and appearance from the multi-view image outputs of these models.


\subsection{3D Reconstruction Models}
Reconstruction models aim to produce 3D representations of an object from a single view or multiple views.
PixelNeRF~\cite{yu2021pixelnerf} achieves single-view 3D reconstruction by projecting the input image features to 3D and applying volume rendering for learning 3D representations.
The recent work, LRM~\cite{hong2023lrm}, greatly boosts the reconstruction quality of PixelNeRF by leveraging a large transformer model and a huge amount of data. 
However, these methods generally synthesize blurry results from unseen viewpoints because they don't model the uncertainty and only use regression losses to train.
This issue can be addressed by using multi-view inputs for the reconstruction model.
For example, by using multi-view images generated by multi-view diffusion models as the input, the 3D reconstruction model, Instant3D~\cite{li2023instant3d}, LGM~\cite{tang2024lgm}, and InstantMesh~\cite{xu2024instantmesh} can reconstruct 3D models from text or image prompts. Specifically, InstantMesh~\cite{xu2024instantmesh} reconstructs 3D meshes with an iso-surface extraction module, \ie, FlexiCubes~\cite{shen2023flexible} representation from multi-view images.

\vspace{-8pt}
\section{Method}
\label{sec:method}
In this section, we introduce \sysname{} -- a novel image-to-3D generative model that achieves both efficient sampling and high-quality generation. More precisely, given a single image of an object and a random noise $\bz$, \sysname{} learns a single-step generator to output 3D representations (we mainly experimented on meshes in this paper) of the object. An illustration of our proposed model is shown in \cref{fig:pipeline} where multi-view images are used as an intermediate representation similar to \cite{li2023instant3d,xu2024instantmesh}. 
We learn our models efficiently using a two-stage distillation approach given pre-trained multi-view diffusion and reconstruction models, where we first learn an efficient multi-view generator based on variational score distillation (VSD, \cref{subsec:gen}), and then finetune our model with a 3D consistent distillation algorithm (\cref{subsec:recon}). The learning pipeline is shown in \cref{fig:learning_pipeline}.

\subsection{Stage I: Multi-view Score Distillation}
\label{subsec:gen}

\topic{Variational Score Distillation (VSD)}
VSD~\cite{wang2023prolificdreamer} is an extension of Score Distillation Sampling (SDS), which was first introduced by DreamFusion~\cite{poole2022dreamfusion} for distilling pre-trained 2D diffusion knowledge into 3D. The core idea of SDS is to match the score function between the output of a learnable parametric image generator and the real data estimated by a pretrained diffusion model. Given a datapoint $\bx = g(\theta)$ generated by the differentiable image generator $g$ parametrized with $\theta$, SDS adds Gaussian noise of level $t$ and turns it into $\bx_t$. It then uses a pre-trained diffusion model with denoising function $\beps_{\phi}(\bx_t; y, t)$ to predict the noise with condition $y$ to optimize $\theta$. 
ProlificDreamer~\cite{wang2023prolificdreamer} proposed VSD to further improve SDS by directly optimizing the distribution of $\theta$ such that the rendering distribution $q(\bx|y)$ with condition $y$ align with the pretrained diffusion model $p(\bx|y)$ by minimizing their KL divergence:
$D_{\mathrm{KL}}(q(\bx|y)||p(\bx|y))$. In practice, this is achieved by learning a separate ``student model'' that estimates the score function of the learned 3D models. The learned score will be used for back-propagation to learn 3D distribution.

\topic{Generative Modeling with VSD}
The original ProlificDreamer parameterized the 3D distribution using a fixed number of particles~\cite{wang2023prolificdreamer}, which, however, does not allow us to draw new samples from the learned distribution. To facilitate learning a 3D generative model that can handle novel scenes, we propose to replace the original parameterization with a learnable generator $\mathcal{G}(\theta)$ that transforms a random Gaussian noise $\eps$ input to a data sample. The training objective of $\mathcal{G}$ is derived as follows:
\begin{align}\label{equ:vsd}
\resizebox{1.0\linewidth}{!}{
$
\nabla_\theta\mathcal{L}_{\text{VSD}} = \mathbb{E}_{t, \eps}\Big[w(t)(\eps_{\text{pre}}(\mathbf{x}_t; y, t) - 
\eps_{\text{stu}}(\mathbf{x}_t; y, t))\frac{\partial \mathcal{G}(\theta, \bz)}{\partial \theta}\Big]
$
}
\end{align}
where $\bx_0 = \mathcal{G}(\theta, \bz)$ is the clean sample of the generator output given noise $\bz \in \mathcal{N}(0, \bI)$ and $\bx_t$ is the noisy version of $\bx_0$, $t$ is the diffusion timestep, $\eps_{\text{pre}}$ and $\eps_{\text{stu}}$ are the predictions of the pretrained diffusion model and the student model respectively. The student model is trained online on the output of $\mathcal{G}$ to estimate the score of the generated samples:
\begin{equation}
    \mathcal{L}_{\text{stu}} = \mathbb{E}_{t, \eps}\lVert\eps_{\text{stu}}(\mathbf{x}_t; y, t))-\eps\rVert_2^2
\end{equation}

\topic{Multi-view Distillation}
Ideally, our goal is to learn a 3D generator that directly maps random noises to 3D representations using VSD, and the 2D renderings of the generator become the input of the 2D diffusion models. 
Here, we can leverage large-scale pre-trained multi-view diffusion models~\cite{shi2023zero123++,long2023wonder3d} as our teacher models to improve the learning of 3D inductive bias.
A natural design would be to parametrize $\mathcal{G}(\theta, \bz)$ with a 3D generator, such as a triplane generator~\cite{chan2022efficient}. However, we found that training a generator from scratch without proper initialization would lead to severe mode collapse, \ie all the samples drawn from the generator will become identical. This observation coincides with the finding in recent work~\cite{luo2024diff, gu2023boot, yin2023one} for the distillation of 2D Diffusion models trained on single-view images. 

To circumvent this problem, we propose to first learn multi-view images as an intermediate representation using VSD. This allows us to use the same architecture and initial parameters as the pretrained model for our generator $\mathcal{G}$ which is essentially a single-step multi-view generator. 
In \sysname{}, we employ Zero123Plus~\cite{shi2023zero123++} as our teacher model because it provides photorealistic and highly consistent 6-view renderings. In contrast to \cite{shi2023zero123++} that uses reference attention~\cite{zhang2022reference} to concatenate the self-attention matrices of the noised condition image, we directly used the self-attention matrices of the clean condition image to preserve the information. As mentioned earlier, we initialize the generator with pretrained Zero123Plus with the additional conversion from $\bv$-prediction to $\bx_0$-prediction. 


\subsection{Stage II: 3D Consistent Distillation}
\label{subsec:recon}
After the multi-view images of the object are obtained, our next step is to estimate the 3D representation of the object from the multi-view images. One potential solution is to 
apply a pretrained 3D reconstruction network $\mathcal{R}$ that takes multi-view images as input and outputs a 3D representation. However, one major drawback of this approach is that the output of the one-step multi-view generator $\mathcal{G}(\theta)$, which is also the input to the reconstruction network $\mathcal{R}$, has low multi-view consistency compared to the ground-truth multi-view images, which causes training-testing mismatch in the reconstruction model, resulting in incorrect geometry in the output 3D reconstruction.

As shown in \cref{fig:learning_pipeline}, we propose a second distillation stage to resolve this inconsistency issue, which finetunes a reconstruction model as part of the generative model. Considering that the multi-view generation of the teacher diffusion model is much more consistent than the learned single-step generator, we can use the 3D representation reconstructed from these images as pseudo ground truth to refine the reconstruction model. Namely, given a condition image $y$ and sampled noise $\bz$, we conduct the deterministic DDIM sampling~\cite{song2020denoising} using Zero123Plus~\cite{shi2023zero123++} to obtain $\bx_{\text{mv}}$. 3D representations are then reconstructed based on the pretrained 3D reconstructor $\mathcal{R}(\bx_{\text{mv}})$. With the reconstructed mesh, we render from random viewpoints to create a set of pseudo ground truth images $\{I^\text{syn}_i(\bz), i=1,...,N\}$. We collect such paired dataset $\mathbf{D} = (\bz$, $\{I^\text{syn}_i(\bz) | i=1,...,N\})$ for each sampled noise $\bz$, and use them for training the final generator which includes our pretrained single-step multi-view generator (described in Sec.~\ref{subsec:gen}) and a pretrained 3D reconstructor~\cite{xu2024instantmesh, li2023instant3d}. Here, $I(\bz)_i$ represents the $i-$th view rendered from the generator given $\bz$.
In practice, we finetune our final generator by minimizing the difference between the renderings of the generator's output and the corresponding pseudo ground truth images $\{I^\text{syn}_i(\bz), i=1,...,N\}$ rendered from the same viewpoint in terms of the RGB loss and LPIPS~\cite{zhang2018unreasonable} loss:
\begin{equation}
\resizebox{1.0\linewidth}{!}{
$
    \mathcal{L}_{\mathrm{3D}} = \mathbb{E}_{\bz, I^\text{syn}(\bz)}\left[\mathcal{L}_{\text{MSE}}(I_{\text{rgb}}(\bz), I_{\text{rgb}}^{\text{syn}}(\bz)) 
    + \lambda\cdot \mathcal{L}_{\text{LPIPS}}(I_{\text{rgb}}(\bz), I_{\text{rgb}}^{\text{syn}}(\bz))\right]
$
}
\end{equation}

Note that the 3D reconstructor in the final generator can be regarded as a refinement module that tackles the multi-view inconsistency issue for 3D reconstruction. Furthermore, this training strategy allows us to go beyond the fixed six-view setting specified in Zero123Plus~\cite{shi2023zero123++} and use the renderings from arbitrary viewpoints for training, which is an important factor for high-quality 3D reconstruction.

\begin{table*}[htbp]
  \caption{Quantitative comparison of novel view synthesis of \sysname{} and the baselines on GSO~\cite{downs2022google} dataset. We report PSNR, SSIM~\cite{wang2004image}, LPIPS~\cite{zhang2018unreasonable} for novel view synthesis, CD and volume IoU for geometry. For the runtime, ``Get 3D" refers to the time that the model predicts the 3D representations (e.g. triplanes, gaussians) from single view inputs, ``3D to mesh" is the time that converts the 3D representations to meshes. We tested all methods on NVIDIA L40.  The best results are \textbf{bolded} and the second best results are \underline{underlined}.}
  \vspace{-8pt}
  \label{tab:headings}
  \centering
  \resizebox{1.\textwidth}{!}{
  \begin{tabular}{@{}lcccccccc}
    \toprule
    \multirow{2}{*}{Method} & \multicolumn{3}{c}{Novel View Synthesis} & \multicolumn{2}{c}{Geometry} & \multicolumn{3}{c}{Runtime}\\
     & PSNR$\uparrow$ & SSIM$\uparrow$ & LPIPS$\downarrow$ & CD$\downarrow$ & vIoU$\uparrow$ & Get 3D (s)$\downarrow$ & 3D to Mesh (s)$\downarrow$ & Total $\downarrow$\\
    \cmidrule{1-1}\cmidrule(lr){2-4}\cmidrule(lr){5-6}\cmidrule(lr){7-9}
    TriplaneGaussian~\cite{zou2023triplane}  & $18.52$ & $0.817$ & $0.191$ & $0.036$ & $0.492$ & $\textbf{0.11}$ & $140.0$ & $140.1$\\
    OpenLRM~\cite{hong2023lrm, openlrm} & $18.15$ & $0.810$ &$0.173$ & $0.035$ & $0.557$ & $0.22$ & $1.06$ & $2.49$ \\
    TripoSR~\cite{tochilkin2024triposr} & $18.33$ & $0.812$ & $0.172$ & $0.033$ & $0.577$ & $\underline{0.16}$ & $1.30$ & $1.46$ \\
    LGM~\cite{tang2024lgm} & $18.11$ & $0.805$ & $0.178$ & $0.038$ & $0.478$ & $1.28$ & $145.9$ & $147.1$\\
    InstantMesh~\cite{xu2024instantmesh} (Our teacher) &  $\underline{19.15}$ & $\underline{0.822}$ & $\textbf{0.152}$ & $\textbf{0.028}$ & $\textbf{0.626}$ & $7.06$ & $\textbf{0.30}$ & $7.36$ \\
    Ours & $\textbf{19.31}$ & $\textbf{0.825}$ & $\underline{0.154}$ & $\underline{0.029}$ & $\underline{0.599}$ & $0.34$ & $\textbf{0.30}$ & $\textbf{0.64}$ \\
  \bottomrule
  \end{tabular}
}
\vspace{-10pt}
\label{tab:3d}
\end{table*}

\section{Experiments}
\label{sec:exp}
\subsection{Implementation Details}
\topic{Datasets}
We train our model on the LVIS subset of the Objaverse~\cite{deitke2023objaverse} dataset, which contains approximately $46,000$ objects. For each scene, we only need images at one viewpoint to be the condition image of Zero123Plus~\cite{shi2023zero123++}.

\topic{Multi-view Score Distillation}
For Stage I training, the multi-view generator, pretrained teacher Zero123Plus model, and student Zero123Plus model as shown in \cref{fig:learning_pipeline} are all initialized with the fine-tuned white background Zero123Plus~\cite{shi2023zero123++} in InstantMesh~\cite{xu2024instantmesh}.
We train the generator and student model on a single NVIDIA L40 GPU for $5,000$ steps.
In each iteration, the generator and student model are updated alternatively. The $t$ for the student model training is randomly sampled from $[0.02, 0.98]$.
We use a fixed guidance scale of 4 for the generator and the pretrained teacher model, and a guidance scale of 1 for the student model. The generator and the student model are both optimized by the Adam optimizer with learning rate 1e-6, and betas ($0.9, 0.999$). We found it is crucial to balance the learning rate of the generator and student model, otherwise the generator will not converge to reasonable results.

\topic{3D Consistent Distillation}
We adopt InstantMesh~\cite{xu2024instantmesh} as our reconstruction network $\mathcal{R}$.
For each condition image, we ran Zero123Plus with deterministic 75-step DDIM scheduler~\cite{song2020denoising} to obtain the pseudo ground truth six views and use it as input of the InstantMesh~\cite{xu2024instantmesh} to inference 3D meshes. Then we render 50 images at random viewpoints to save them for Stage II training. In Stage II, we use a learning rate of 1e-6 and a batch size of 8 to train 10 epochs.

\topic{Inference}
The whole pipeline of \sysname{} takes about 0.64s for each scene to generate 3D meshes on a single NVIDIA L40 GPU, including 0.28s for multi-view image generation and 0.06s for flexicube reconstruction and 0.30s for mesh extraction. It consumes about 10 GB of GPU memory during inference.


\begin{figure*}[tp]
    \centering
    \includegraphics[width=\linewidth]{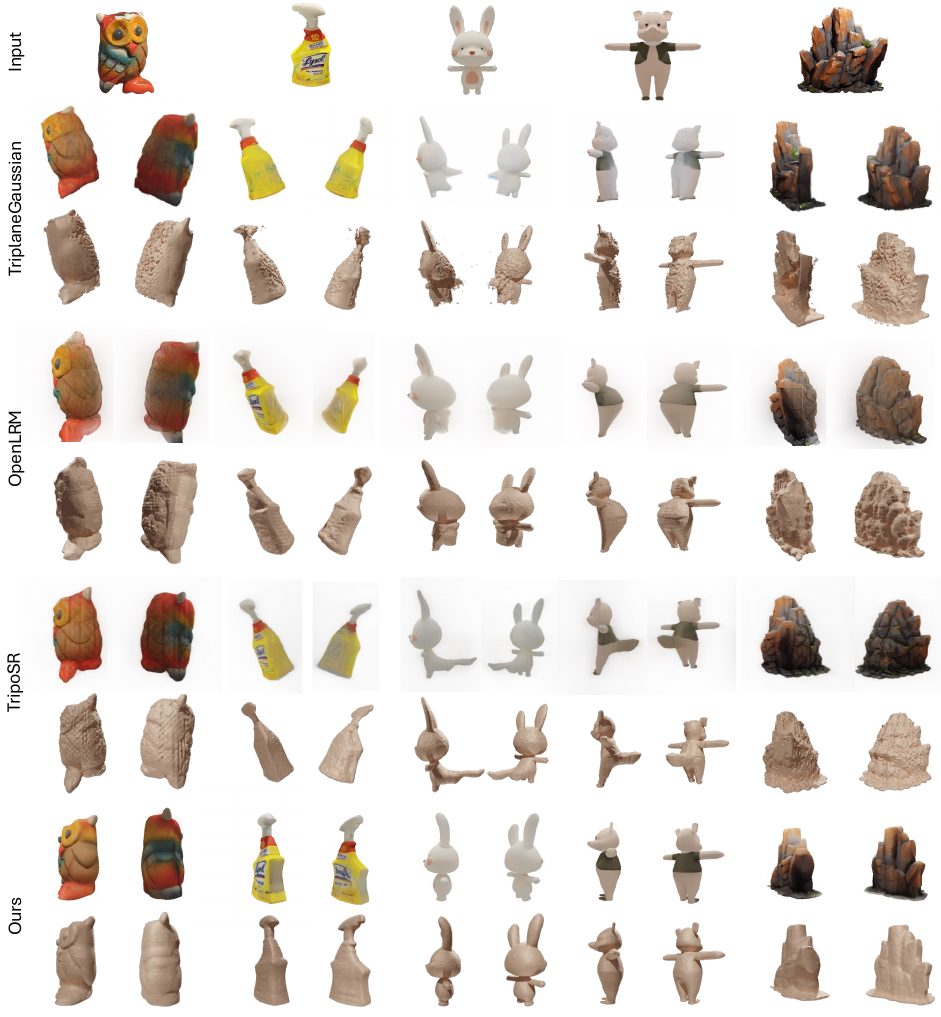}
    \caption{Qualitative comparison against baseline methods. \sysname{} outperforms the baselines, especially from the unseen views. For each method, the first row and second row are the texture and geometry renderings respectively.}
    \vspace{-15pt}
    \label{fig:comparison}
\end{figure*}

\vspace{-5pt}

\subsection{Experiment Protocol}
\vspace{-5pt}
\label{subsec:exp-proc}
\topic{Evaluation Dataset and Metrics} Following prior works~\cite{liu2023syncdreamer, liu2023zero, liu2024one}, we adopt the Google Scanned Object (GSO)~\cite{downs2022google} dataset to perform the quantitative comparison of all the methods. We use the same randomly sampled 30 objects ranging from daily objects to animals in SyncDreamer~\cite{liu2023syncdreamer}. For each object, we render an image with a size of $512 \times 512$ as the input view with zero elevation and render another two sets for evaluation: the first set consists of 6 images from the same viewpoint as in Zero123Plus~\cite{shi2023zero123++}, the second consists of evenly sampled 15 images around the object with zero elevation. For novel view synthesis, we employ commonly used metrics for evaluation, including PSNR, SSIM~\cite{wang2004image} and LPIPS~\cite{zhang2018unreasonable}. For geometry evaluation, we report chamfer distance (CD) and Volume IoU (vIoU).

\topic{Baselines}
We mainly compare with recent methods that focus on feed-forward 3D generation, including LRM~\cite{hong2023lrm}, TriplaneGaussian~\cite{zou2023triplane} and TripoSR~\cite{tochilkin2024triposr}. 
We use the community version OpenLRM~\cite{openlrm} for LRM comparison since the original model is not publicly available. We also include LGM~\cite{tang2024lgm}, which generates 3D Gaussians from multi-view images in seconds.


%

\vspace{-5pt}
\subsection{Results}
\vspace{-8pt}
\begin{figure}[htbp]
    \centering
    \includegraphics[width=\linewidth]{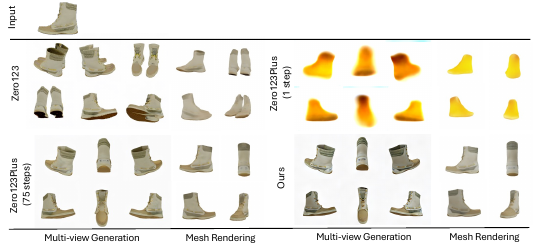}
    \caption{Comparison of \sysname{} with the baselines that use different multi-view image generation methods and then reconstruct 3D meshes. Our one-step multi-view generator produces much better results than Zero123~\cite{liu2023zero} and is comparable to Zero123Plus~\cite{shi2023zero123++} 75-step sampling, leading to better 3D renderings.}
    \label{fig:mvresults}
    \vspace{-5pt}
\end{figure}

\topic{Qualitative Comparison}
\cref{fig:comparison} demonstrates the renderings of \sysname{} and other baselines. We urge readers to view our supplemental video to judge the multi-view consistency of the results. Due to the reconstruction nature, the baseline methods fail to generate reasonable textures at unseen viewpoints, producing incorrect geometry and blurry renderings. Our method handles the uncertainty through multi-view image generation, so even at the back viewpoints, we can synthesize details that are highly consistent with the input image. From the geometry renderings, we can see that our method also generates consistent and smooth geometry with the input image at the back viewpoints, outperforming other baselines. Video comparisons with baselines can be found in the supplementary.

\cref{fig:mvresults} further shows comparisons with other methods that also generate multi-view images first and then reconstruct 3D meshes with InstantMesh. It can be seen from \cref{fig:mvresults} that the output of Zero123~\cite{liu2023zero} is not consistent across different viewpoints because each view is generated separately, \eg the shoes are in a pair in one view but not in another. Therefore, the rendered images from 3D meshes tend to be blurry and lack of details. Our method generates multi-view images that are much more consistent and look similar to 75-step sampling of Zero123Plus~\cite{shi2023zero123++}. The high-quality multi-view generation provides a basis for the 3D generation stage.

\topic{Quantitative Comparison}
The quantitative comparison is shown in \tabref{tab:3d}. Our method achieves the best results in novel view synthesis and geometry metrics. The improvement results from the uncertainty handling of our approach in the occluded regions. In contrast, TriplaneGaussian~\cite{zou2023triplane}, LRM~\cite{hong2023lrm} and TripoSR~\cite{tochilkin2024triposr} cannot produce sharp predictions for unseen parts, as they are deterministic models. 
In terms of runtime, thanks to the Flexicubes technique, our method achieves a higher rendering speed and more efficient conversion to meshes than other methods. TriplaneGaussian achieves highly efficient rendering with 3D Gaussians, but requires a slow mesh conversion process with Point-E~\cite{nichol2022point}. OpenLRM and TripoSR render much slower due to triplane representation. Our method also achieves comparable performance compared with our teacher model InstantMesh~\cite{xu2024instantmesh} and better results than LGM~\cite{tang2024lgm}, while being much faster.

\begin{figure}[tp]
    \centering
    \includegraphics[width=1.0\linewidth]{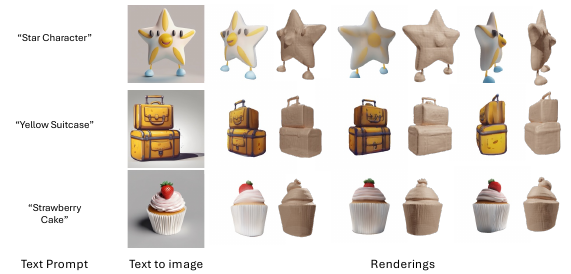}
    \caption{3D generation given text prompts.}
    \label{fig:text23d}
    \vspace{-20pt}
\end{figure}

\topic{Text-to-image-to-3D Generation} Our method can also be combined with text-to-image diffusion models for 3D generation from text prompts. We first use SD-XL~\cite{podell2023sdxl} to generate images and then run \sysname{} to synthesize 3D. The results are shown in \cref{fig:text23d}.

\begin{table}
  \caption{Quantataively results on the renderings and geometry across different settings. We report PSNR, SSIM~\cite{wang2004image}, LPIPS~\cite{zhang2018unreasonable}, CD and vIoU on the GSO~\cite{downs2022google} dataset.}
  \vspace{-5pt}
  \centering
\resizebox{1.0\linewidth}{!}{
  \begin{tabular}{@{}lccccc}
    \toprule
    Method & PSNR$\uparrow$ & SSIM$\uparrow$ & LPIPS$\downarrow$ & CD$\downarrow$ & vIoU$\uparrow$\\
    \midrule
    1-step Zero123Plus with InstantMesh~\cite{shi2023zero123++}  & $14.79$ & $0.791$ & $0.262$ & $0.052$ & $0.462$ \\
    Ours w/o Stage II & $18.75$ & $0.812$ & $0.157$ & \textbf{0.029} & 0.585 \\
    Ours &  $\mathbf{19.30}$ & $\mathbf{0.825}$ & $\mathbf{0.154}$ & \textbf{0.029} & \textbf{0.599} \\
  \bottomrule
  \end{tabular}
}
  \label{tab:ablation}
  \vspace{-10pt}
\end{table}

\subsection{Ablation Study}

\begin{figure}[tp]
    \centering
    \includegraphics[width=1.0\linewidth]{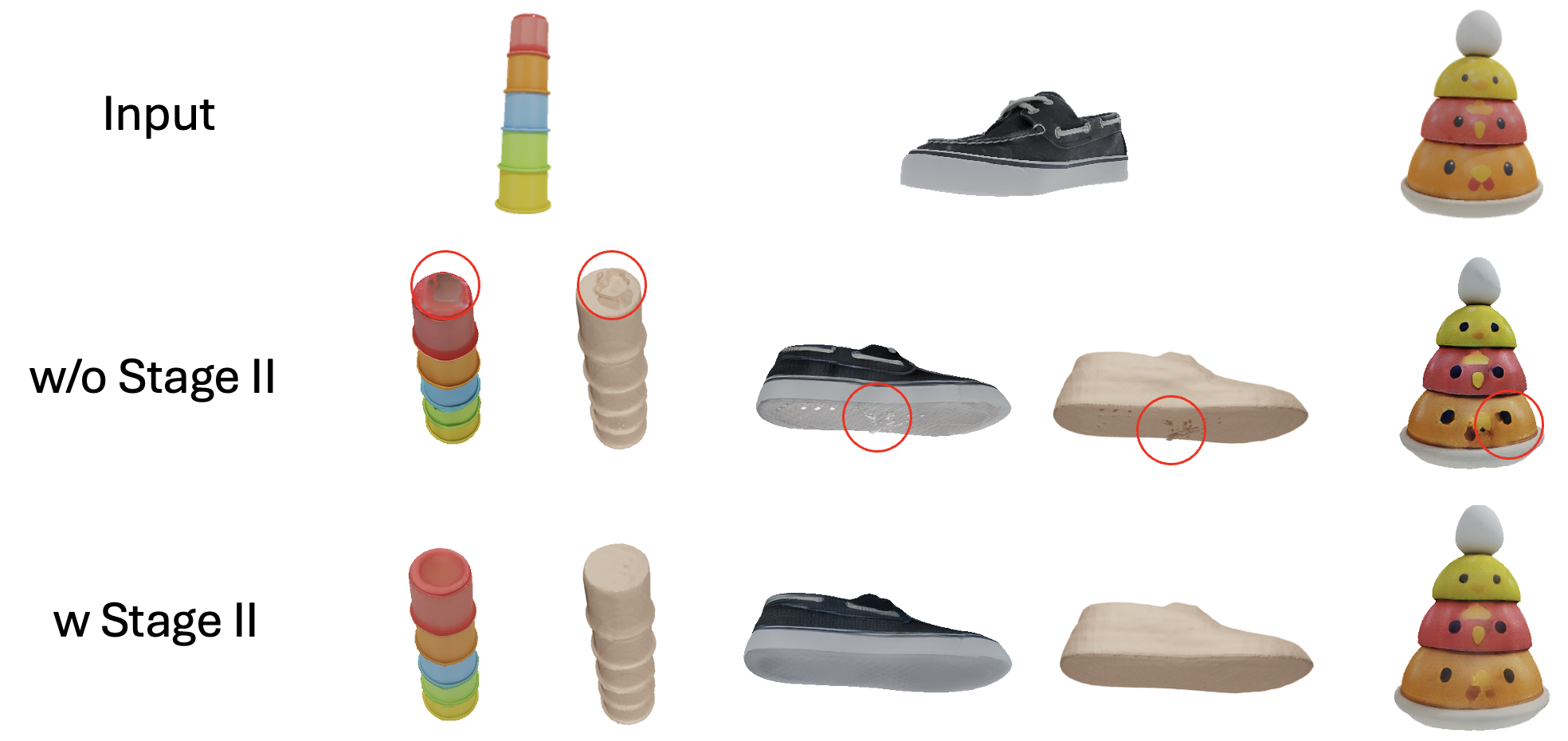}
    \vspace{-20pt}
    \caption{The Stage-II training alleviates the view inconsistency issue in the multi-view diffusion outputs, resulting in higher-quality results with less bad geometry and overexposure.}
    \label{fig:ablation}
    \vspace{-15pt}
\end{figure}

\topic{Quality of Multi-View Generator} We evaluated the generated 6 views of our multi-view image generator on the same GSO~\cite{downs2022google} objects we used in \cref{subsec:exp-proc}. Results show that our multi-view generator after VSD distillation achieves PSNR/SSIM/LPIPS of $17.28/0.784/0.197$, which is close to the results of 75-step sampling Zero123Plus~\cite{shi2023zero123++} ($18.00/0.790/0.190$). From \cref{fig:mvresults}, we can see that 1-step inference in the original Zero123Plus cannot produce reasonable results, while our method is able to generate sharp multi-view images and high-quality 3D renderings.

\topic{Effectiveness of Stage-II} We compare the generation results before and after stage-II training in \cref{fig:ablation}. We can see from the results that after the stage-II training, the incorrect geometry on the object has gone. This is because the Stage-II training distills a more robust 3D generator that can handle the inconsistency of input images benefiting from the multi-view supervision from the pseudo ground truth images. Also, the overexposure problem in the generated multi-view images from stage-II is also resolved (rightmost example in \cref{fig:ablation}). For \tabref{tab:ablation}, we can also see that both the novel view synthesis metrics and geometry metrics are improved after stage-II.

\section{Conclusion and Future Work}
In this work, we present \sysname{}, a generative framework for 3D content generation. 
We found that directly learning a 3D generative model that generalizes well involves learning from massive 3D data. Therefore, we use the intermediate representation and employ a multi-view image generation and reconstruction framework. The uncertainty of 3D generation is well addressed in the multi-view image diffusion stage that enjoys the rich prior of pretrained 2D image diffusion models. Then, 3D can be obtained through multi-view reconstruction. We further jointly learn the multi-view image generator and reconstructor to improve the 3D consistency. The whole pipeline is feed-forward and requires less than one second.

Our approach still has some limitations. First, our training process involves two stages, including distilling a multi-view image diffusion model and leveraging it to learn a reconstruction model.
Second, the results of our work are bounded by the multi-step sampling results of the multi-view diffusion models, which might still not be as consistent as renderings of 3D representations. The first stage might also influence the diversity of the generator.
Future work can consider learning a one-step 3D generative model that can produce 3D representations directly, either by training from scratch or distilling a 3D diffusion model.

{
    \small
    \bibliographystyle{ieeenat_fullname}
    \bibliography{main}

\begin{thebibliography}{88}
\providecommand{\natexlab}[1]{#1}
\providecommand{\url}[1]{\texttt{#1}}
\expandafter\ifx\csname urlstyle\endcsname\relax
  \providecommand{\doi}[1]{doi: #1}\else
  \providecommand{\doi}{doi: \begingroup \urlstyle{rm}\Url}\fi

\bibitem[Bao et~al.(2021)Bao, Li, Zhu, and Zhang]{bao2022analytic}
Fan Bao, Chongxuan Li, Jun Zhu, and Bo Zhang.
\newblock Analytic-dpm: an analytic estimate of the optimal reverse variance in diffusion probabilistic models.
\newblock \emph{ICLR}, 2021.

\bibitem[Cao et~al.(2023)Cao, Cao, Han, Shan, and Wong]{cao2023dreamavatar}
Yukang Cao, Yan-Pei Cao, Kai Han, Ying Shan, and Kwan-Yee~K Wong.
\newblock Dreamavatar: Text-and-shape guided 3d human avatar generation via diffusion models.
\newblock \emph{arXiv preprint arXiv:2304.00916}, 2023.

\bibitem[Chan et~al.(2022)Chan, Lin, Chan, Nagano, Pan, De~Mello, Gallo, Guibas, Tremblay, Khamis, et~al.]{chan2022efficient}
Eric~R Chan, Connor~Z Lin, Matthew~A Chan, Koki Nagano, Boxiao Pan, Shalini De~Mello, Orazio Gallo, Leonidas~J Guibas, Jonathan Tremblay, Sameh Khamis, et~al.
\newblock Efficient geometry-aware 3d generative adversarial networks.
\newblock In \emph{CVPR}, pages 16123--16133, 2022.

\bibitem[Chen et~al.(2023{\natexlab{a}})Chen, Chen, Jiao, and Jia]{chen2023fantasia3d}
Rui Chen, Yongwei Chen, Ningxin Jiao, and Kui Jia.
\newblock Fantasia3d: Disentangling geometry and appearance for high-quality text-to-3d content creation.
\newblock \emph{ICCV}, 2023{\natexlab{a}}.

\bibitem[Chen et~al.(2023{\natexlab{b}})Chen, Wang, and Liu]{chen2023text}
Zilong Chen, Feng Wang, and Huaping Liu.
\newblock Text-to-3d using gaussian splatting.
\newblock \emph{arXiv preprint arXiv:2309.16585}, 2023{\natexlab{b}}.

\bibitem[Chou et~al.(2023)Chou, Bahat, and Heide]{chou2023diffusion}
Gene Chou, Yuval Bahat, and Felix Heide.
\newblock Diffusion-sdf: Conditional generative modeling of signed distance functions.
\newblock In \emph{CVPR}, pages 2262--2272, 2023.

\bibitem[Deitke et~al.(2023)Deitke, Liu, Wallingford, Ngo, Michel, Kusupati, Fan, Laforte, Voleti, Gadre, et~al.]{deitke2023objaverse}
Matt Deitke, Ruoshi Liu, Matthew Wallingford, Huong Ngo, Oscar Michel, Aditya Kusupati, Alan Fan, Christian Laforte, Vikram Voleti, Samir~Yitzhak Gadre, et~al.
\newblock Objaverse-xl: A universe of 10m+ 3d objects.
\newblock \emph{arXiv preprint arXiv:2307.05663}, 2023.

\bibitem[Dhariwal and Nichol(2021)]{dhariwal2021diffusion}
Prafulla Dhariwal and Alexander Nichol.
\newblock Diffusion models beat gans on image synthesis.
\newblock \emph{NeurIPS}, 34:\penalty0 8780--8794, 2021.

\bibitem[Downs et~al.(2022)Downs, Francis, Koenig, Kinman, Hickman, Reymann, McHugh, and Vanhoucke]{downs2022google}
Laura Downs, Anthony Francis, Nate Koenig, Brandon Kinman, Ryan Hickman, Krista Reymann, Thomas~B McHugh, and Vincent Vanhoucke.
\newblock Google scanned objects: A high-quality dataset of 3d scanned household items.
\newblock In \emph{2022 International Conference on Robotics and Automation (ICRA)}, pages 2553--2560. IEEE, 2022.

\bibitem[Gong et~al.(2023)Gong, Li, Feng, Wu, and Kong]{gong2022diffuseq}
Shansan Gong, Mukai Li, Jiangtao Feng, Zhiyong Wu, and LingPeng Kong.
\newblock Diffuseq: Sequence to sequence text generation with diffusion models.
\newblock \emph{ICLR}, 2023.

\bibitem[Gu et~al.(2023)Gu, Trevithick, Lin, Susskind, Theobalt, Liu, and Ramamoorthi]{gu2023nerfdiff}
Jiatao Gu, Alex Trevithick, Kai-En Lin, Joshua~M Susskind, Christian Theobalt, Lingjie Liu, and Ravi Ramamoorthi.
\newblock Nerfdiff: Single-image view synthesis with nerf-guided distillation from 3d-aware diffusion.
\newblock In \emph{ICML}, pages 11808--11826. PMLR, 2023.

\bibitem[Gu et~al.(2024)Gu, Wang, Zhai, Zhang, Liu, and Susskind]{gu2023boot}
Jiatao Gu, Chen Wang, Shuangfei Zhai, Yizhe Zhang, Lingjie Liu, and Joshua~M Susskind.
\newblock Data-free distillation of diffusion models with bootstrapping.
\newblock In \emph{ICML}, 2024.

\bibitem[He and Wang(2023)]{openlrm}
Zexin He and Tengfei Wang.
\newblock Openlrm: Open-source large reconstruction models.
\newblock \url{https://github.com/3DTopia/OpenLRM}, 2023.

\bibitem[Hinton et~al.(2015)Hinton, Vinyals, and Dean]{hinton2015distilling}
Geoffrey Hinton, Oriol Vinyals, and Jeff Dean.
\newblock Distilling the knowledge in a neural network.
\newblock \emph{arXiv preprint arXiv:1503.02531}, 2015.

\bibitem[Ho et~al.(2020)Ho, Jain, and Abbeel]{ho2020denoising}
Jonathan Ho, Ajay Jain, and Pieter Abbeel.
\newblock Denoising diffusion probabilistic models.
\newblock \emph{NeurIPS}, 33:\penalty0 6840--6851, 2020.

\bibitem[Ho et~al.(2022{\natexlab{a}})Ho, Chan, Saharia, Whang, Gao, Gritsenko, Kingma, Poole, Norouzi, Fleet, et~al.]{ho2022imagen}
Jonathan Ho, William Chan, Chitwan Saharia, Jay Whang, Ruiqi Gao, Alexey Gritsenko, Diederik~P Kingma, Ben Poole, Mohammad Norouzi, David~J Fleet, et~al.
\newblock Imagen video: High definition video generation with diffusion models.
\newblock \emph{arXiv preprint arXiv:2210.02303}, 2022{\natexlab{a}}.

\bibitem[Ho et~al.(2022{\natexlab{b}})Ho, Salimans, Gritsenko, Chan, Norouzi, and Fleet]{ho2022video}
Jonathan Ho, Tim Salimans, Alexey~A. Gritsenko, William Chan, Mohammad Norouzi, and David~J. Fleet.
\newblock Video diffusion models.
\newblock In \emph{ICLR Workshop on Deep Generative Models for Highly Structured Data}, 2022{\natexlab{b}}.

\bibitem[H{\"o}llein et~al.(2023)H{\"o}llein, Cao, Owens, Johnson, and Nie{\ss}ner]{hollein2023text2room}
Lukas H{\"o}llein, Ang Cao, Andrew Owens, Justin Johnson, and Matthias Nie{\ss}ner.
\newblock Text2room: Extracting textured 3d meshes from 2d text-to-image models.
\newblock \emph{ICCV}, 2023.

\bibitem[Hong et~al.(2024)Hong, Zhang, Gu, Bi, Zhou, Liu, Liu, Sunkavalli, Bui, and Tan]{hong2023lrm}
Yicong Hong, Kai Zhang, Jiuxiang Gu, Sai Bi, Yang Zhou, Difan Liu, Feng Liu, Kalyan Sunkavalli, Trung Bui, and Hao Tan.
\newblock Lrm: Large reconstruction model for single image to 3d.
\newblock \emph{ICLR}, 2024.

\bibitem[Jakab et~al.(2023)Jakab, Li, Wu, Rupprecht, and Vedaldi]{jakab2023farm3d}
Tomas Jakab, Ruining Li, Shangzhe Wu, Christian Rupprecht, and Andrea Vedaldi.
\newblock Farm3d: Learning articulated 3d animals by distilling 2d diffusion.
\newblock \emph{arXiv preprint arXiv:2304.10535}, 2023.

\bibitem[Jun and Nichol(2023)]{jun2023shap}
Heewoo Jun and Alex Nichol.
\newblock Shap-e: Generating conditional 3d implicit functions.
\newblock \emph{arXiv preprint arXiv:2305.02463}, 2023.

\bibitem[Karras et~al.(2022)Karras, Aittala, Aila, and Laine]{karras2022elucidating}
Tero Karras, Miika Aittala, Timo Aila, and Samuli Laine.
\newblock Elucidating the design space of diffusion-based generative models.
\newblock \emph{NeurIPS}, 35:\penalty0 26565--26577, 2022.

\bibitem[Kerbl et~al.(2023)Kerbl, Kopanas, Leimk{\"u}hler, and Drettakis]{kerbl20233dgs}
Bernhard Kerbl, Georgios Kopanas, Thomas Leimk{\"u}hler, and George Drettakis.
\newblock 3d gaussian splatting for real-time radiance field rendering.
\newblock \emph{ACM TOG}, 42\penalty0 (4), 2023.

\bibitem[Kwak et~al.(2023)Kwak, Dong, Jin, Ko, Mahajan, and Yi]{kwak2023vivid}
Jeong-gi Kwak, Erqun Dong, Yuhe Jin, Hanseok Ko, Shweta Mahajan, and Kwang~Moo Yi.
\newblock Vivid-1-to-3: Novel view synthesis with video diffusion models.
\newblock \emph{arXiv preprint arXiv:2312.01305}, 2023.

\bibitem[Li et~al.(2024)Li, Tan, Zhang, Xu, Luan, Xu, Hong, Sunkavalli, Shakhnarovich, and Bi]{li2023instant3d}
Jiahao Li, Hao Tan, Kai Zhang, Zexiang Xu, Fujun Luan, Yinghao Xu, Yicong Hong, Kalyan Sunkavalli, Greg Shakhnarovich, and Sai Bi.
\newblock Instant3d: Fast text-to-3d with sparse-view generation and large reconstruction model.
\newblock \emph{ICLR}, 2024.

\bibitem[Li et~al.(2023)Li, Dou, Shi, Lei, Chen, Zhang, Zhou, and Ni]{li2023focaldreamer}
Yuhan Li, Yishun Dou, Yue Shi, Yu Lei, Xuanhong Chen, Yi Zhang, Peng Zhou, and Bingbing Ni.
\newblock Focaldreamer: Text-driven 3d editing via focal-fusion assembly.
\newblock \emph{arXiv preprint arXiv:2308.10608}, 2023.

\bibitem[Liang et~al.(2023)Liang, Yang, Lin, Li, Xu, and Chen]{liang2023luciddreamer}
Yixun Liang, Xin Yang, Jiantao Lin, Haodong Li, Xiaogang Xu, and Yingcong Chen.
\newblock Luciddreamer: Towards high-fidelity text-to-3d generation via interval score matching.
\newblock \emph{arXiv preprint arXiv:2311.11284}, 2023.

\bibitem[Liao et~al.(2023)Liao, Yi, Xiu, Tang, Huang, Thies, and Black]{liao2023tada}
Tingting Liao, Hongwei Yi, Yuliang Xiu, Jiaxaing Tang, Yangyi Huang, Justus Thies, and Michael~J Black.
\newblock Tada! text to animatable digital avatars.
\newblock \emph{arXiv preprint arXiv:2308.10899}, 2023.

\bibitem[Lin et~al.(2023)Lin, Gao, Tang, Takikawa, Zeng, Huang, Kreis, Fidler, Liu, and Lin]{lin2023magic3d}
Chen-Hsuan Lin, Jun Gao, Luming Tang, Towaki Takikawa, Xiaohui Zeng, Xun Huang, Karsten Kreis, Sanja Fidler, Ming-Yu Liu, and Tsung-Yi Lin.
\newblock Magic3d: High-resolution text-to-3d content creation.
\newblock In \emph{CVPR}, pages 300--309, 2023.

\bibitem[Liu et~al.(2024{\natexlab{a}})Liu, Xu, Jin, Chen, Varma~T, Xu, and Su]{liu2024one}
Minghua Liu, Chao Xu, Haian Jin, Linghao Chen, Mukund Varma~T, Zexiang Xu, and Hao Su.
\newblock One-2-3-45: Any single image to 3d mesh in 45 seconds without per-shape optimization.
\newblock \emph{NeurIPS}, 36, 2024{\natexlab{a}}.

\bibitem[Liu et~al.(2023)Liu, Wu, Van~Hoorick, Tokmakov, Zakharov, and Vondrick]{liu2023zero}
Ruoshi Liu, Rundi Wu, Basile Van~Hoorick, Pavel Tokmakov, Sergey Zakharov, and Carl Vondrick.
\newblock Zero-1-to-3: Zero-shot one image to 3d object.
\newblock In \emph{ICCV}, pages 9298--9309, 2023.

\bibitem[Liu et~al.(2024{\natexlab{b}})Liu, Lin, Zeng, Long, Liu, Komura, and Wang]{liu2023syncdreamer}
Yuan Liu, Cheng Lin, Zijiao Zeng, Xiaoxiao Long, Lingjie Liu, Taku Komura, and Wenping Wang.
\newblock Syncdreamer: Generating multiview-consistent images from a single-view image.
\newblock \emph{ICLR}, 2024{\natexlab{b}}.

\bibitem[Liu et~al.()Liu, Guo, Voleti, Shao, Chen, Luo, Zou, Wang, Laforte, Cao, et~al.]{liuthreestudio}
Ying-Tian Liu, Yuan-Chen Guo, Vikram Voleti, Ruizhi Shao, Chia-Hao Chen, Guan Luo, Zixin Zou, Chen Wang, Christian Laforte, Yan-Pei Cao, et~al.
\newblock threestudio: a modular framework for diffusion-guided 3d generation.

\bibitem[Long et~al.(2023)Long, Guo, Lin, Liu, Dou, Liu, Ma, Zhang, Habermann, Theobalt, et~al.]{long2023wonder3d}
Xiaoxiao Long, Yuan-Chen Guo, Cheng Lin, Yuan Liu, Zhiyang Dou, Lingjie Liu, Yuexin Ma, Song-Hai Zhang, Marc Habermann, Christian Theobalt, et~al.
\newblock Wonder3d: Single image to 3d using cross-domain diffusion.
\newblock \emph{arXiv preprint arXiv:2310.15008}, 2023.

\bibitem[Lorraine et~al.(2023)Lorraine, Xie, Zeng, Lin, Takikawa, Sharp, Lin, Liu, Fidler, and Lucas]{lorraine2023att3d}
Jonathan Lorraine, Kevin Xie, Xiaohui Zeng, Chen-Hsuan Lin, Towaki Takikawa, Nicholas Sharp, Tsung-Yi Lin, Ming-Yu Liu, Sanja Fidler, and James Lucas.
\newblock Att3d: Amortized text-to-3d object synthesis.
\newblock \emph{arXiv preprint arXiv:2306.07349}, 2023.

\bibitem[Lu et~al.(2022)Lu, Zhou, Bao, Chen, Li, and Zhu]{lu2022dpm}
Cheng Lu, Yuhao Zhou, Fan Bao, Jianfei Chen, Chongxuan Li, and Jun Zhu.
\newblock Dpm-solver: A fast ode solver for diffusion probabilistic model sampling in around 10 steps.
\newblock \emph{NeurIPS}, 35:\penalty0 5775--5787, 2022.

\bibitem[Lu et~al.(2023)Lu, Zhang, Li, Fang, McKinnon, Tsin, Quan, Cao, and Yao]{lu2023direct2}
Yuanxun Lu, Jingyang Zhang, Shiwei Li, Tian Fang, David McKinnon, Yanghai Tsin, Long Quan, Xun Cao, and Yao Yao.
\newblock Direct2. 5: Diverse text-to-3d generation via multi-view 2.5 d diffusion.
\newblock \emph{arXiv preprint arXiv:2311.15980}, 2023.

\bibitem[Luo and Hu(2021)]{luo2021diffusion}
Shitong Luo and Wei Hu.
\newblock Diffusion probabilistic models for 3d point cloud generation.
\newblock In \emph{CVPR}, pages 2837--2845, 2021.

\bibitem[Luo et~al.(2023)Luo, Tan, Huang, Li, and Zhao]{luo2023latent}
Simian Luo, Yiqin Tan, Longbo Huang, Jian Li, and Hang Zhao.
\newblock Latent consistency models: Synthesizing high-resolution images with few-step inference.
\newblock \emph{arXiv preprint arXiv:2310.04378}, 2023.

\bibitem[Luo et~al.(2024)Luo, Hu, Zhang, Sun, Li, and Zhang]{luo2024diff}
Weijian Luo, Tianyang Hu, Shifeng Zhang, Jiacheng Sun, Zhenguo Li, and Zhihua Zhang.
\newblock Diff-instruct: A universal approach for transferring knowledge from pre-trained diffusion models.
\newblock \emph{NeurIPS}, 36, 2024.

\bibitem[Metzer et~al.(2023)Metzer, Richardson, Patashnik, Giryes, and Cohen-Or]{metzer2023latent}
Gal Metzer, Elad Richardson, Or Patashnik, Raja Giryes, and Daniel Cohen-Or.
\newblock Latent-nerf for shape-guided generation of 3d shapes and textures.
\newblock In \emph{CVPR}, pages 12663--12673, 2023.

\bibitem[Mildenhall et~al.(2021)Mildenhall, Srinivasan, Tancik, Barron, Ramamoorthi, and Ng]{mildenhall2021nerf}
Ben Mildenhall, Pratul~P Srinivasan, Matthew Tancik, Jonathan~T Barron, Ravi Ramamoorthi, and Ren Ng.
\newblock Nerf: Representing scenes as neural radiance fields for view synthesis.
\newblock \emph{Communications of the ACM}, 65\penalty0 (1):\penalty0 99--106, 2021.

\bibitem[M{\"u}ller et~al.(2023)M{\"u}ller, Siddiqui, Porzi, Bulo, Kontschieder, and Nie{\ss}ner]{muller2023diffrf}
Norman M{\"u}ller, Yawar Siddiqui, Lorenzo Porzi, Samuel~Rota Bulo, Peter Kontschieder, and Matthias Nie{\ss}ner.
\newblock Diffrf: Rendering-guided 3d radiance field diffusion.
\newblock In \emph{CVPR}, pages 4328--4338, 2023.

\bibitem[M{\"u}ller et~al.(2022)M{\"u}ller, Evans, Schied, and Keller]{muller2022instant}
Thomas M{\"u}ller, Alex Evans, Christoph Schied, and Alexander Keller.
\newblock Instant neural graphics primitives with a multiresolution hash encoding.
\newblock \emph{ACM TOG}, 41\penalty0 (4):\penalty0 1--15, 2022.

\bibitem[Nichol et~al.(2022)Nichol, Jun, Dhariwal, Mishkin, and Chen]{nichol2022point}
Alex Nichol, Heewoo Jun, Prafulla Dhariwal, Pamela Mishkin, and Mark Chen.
\newblock Point-e: A system for generating 3d point clouds from complex prompts.
\newblock \emph{arXiv preprint arXiv:2212.08751}, 2022.

\bibitem[Po and Wetzstein(2023)]{po2023compositional}
Ryan Po and Gordon Wetzstein.
\newblock Compositional 3d scene generation using locally conditioned diffusion.
\newblock \emph{arXiv preprint arXiv:2303.12218}, 2023.

\bibitem[Podell et~al.(2023)Podell, English, Lacey, Blattmann, Dockhorn, M{\"u}ller, Penna, and Rombach]{podell2023sdxl}
Dustin Podell, Zion English, Kyle Lacey, Andreas Blattmann, Tim Dockhorn, Jonas M{\"u}ller, Joe Penna, and Robin Rombach.
\newblock Sdxl: Improving latent diffusion models for high-resolution image synthesis.
\newblock \emph{arXiv preprint arXiv:2307.01952}, 2023.

\bibitem[Poole et~al.(2023)Poole, Jain, Barron, and Mildenhall]{poole2022dreamfusion}
Ben Poole, Ajay Jain, Jonathan~T Barron, and Ben Mildenhall.
\newblock Dreamfusion: Text-to-3d using 2d diffusion.
\newblock \emph{ICLR}, 2023.

\bibitem[Qian et~al.(2024{\natexlab{a}})Qian, Cao, Siarohin, Kant, Wang, Vasilkovsky, Lee, Fang, Skorokhodov, Zhuang, et~al.]{qian2024atom}
Guocheng Qian, Junli Cao, Aliaksandr Siarohin, Yash Kant, Chaoyang Wang, Michael Vasilkovsky, Hsin-Ying Lee, Yuwei Fang, Ivan Skorokhodov, Peiye Zhuang, et~al.
\newblock Atom: Amortized text-to-mesh using 2d diffusion.
\newblock \emph{arXiv preprint arXiv:2402.00867}, 2024{\natexlab{a}}.

\bibitem[Qian et~al.(2024{\natexlab{b}})Qian, Mai, Hamdi, Ren, Siarohin, Li, Lee, Skorokhodov, Wonka, Tulyakov, et~al.]{qian2023magic123}
Guocheng Qian, Jinjie Mai, Abdullah Hamdi, Jian Ren, Aliaksandr Siarohin, Bing Li, Hsin-Ying Lee, Ivan Skorokhodov, Peter Wonka, Sergey Tulyakov, et~al.
\newblock Magic123: One image to high-quality 3d object generation using both 2d and 3d diffusion priors.
\newblock \emph{ICLR}, 2024{\natexlab{b}}.

\bibitem[Rombach et~al.(2022)Rombach, Blattmann, Lorenz, Esser, and Ommer]{rombach2022high}
Robin Rombach, Andreas Blattmann, Dominik Lorenz, Patrick Esser, and Bj{\"o}rn Ommer.
\newblock High-resolution image synthesis with latent diffusion models.
\newblock In \emph{CVPR}, pages 10684--10695, 2022.

\bibitem[Saharia et~al.(2022)Saharia, Chan, Saxena, Li, Whang, Denton, Ghasemipour, Gontijo~Lopes, Karagol~Ayan, Salimans, et~al.]{saharia2022photorealistic}
Chitwan Saharia, William Chan, Saurabh Saxena, Lala Li, Jay Whang, Emily~L Denton, Kamyar Ghasemipour, Raphael Gontijo~Lopes, Burcu Karagol~Ayan, Tim Salimans, et~al.
\newblock Photorealistic text-to-image diffusion models with deep language understanding.
\newblock \emph{NeurIPS}, 35:\penalty0 36479--36494, 2022.

\bibitem[Salimans and Ho(2022)]{salimans2022progressive}
Tim Salimans and Jonathan Ho.
\newblock Progressive distillation for fast sampling of diffusion models.
\newblock \emph{ICLR}, 2022.

\bibitem[Sargent et~al.(2023)Sargent, Li, Shah, Herrmann, Yu, Zhang, Chan, Lagun, Fei-Fei, Sun, et~al.]{sargent2023zeronvs}
Kyle Sargent, Zizhang Li, Tanmay Shah, Charles Herrmann, Hong-Xing Yu, Yunzhi Zhang, Eric~Ryan Chan, Dmitry Lagun, Li Fei-Fei, Deqing Sun, et~al.
\newblock Zeronvs: Zero-shot 360-degree view synthesis from a single real image.
\newblock \emph{arXiv preprint arXiv:2310.17994}, 2023.

\bibitem[Sauer et~al.(2023)Sauer, Lorenz, Blattmann, and Rombach]{sauer2023adversarial}
Axel Sauer, Dominik Lorenz, Andreas Blattmann, and Robin Rombach.
\newblock Adversarial diffusion distillation.
\newblock \emph{arXiv preprint arXiv:2311.17042}, 2023.

\bibitem[Shen et~al.(2023)Shen, Munkberg, Hasselgren, Yin, Wang, Chen, Gojcic, Fidler, Sharp, and Gao]{shen2023flexible}
Tianchang Shen, Jacob Munkberg, Jon Hasselgren, Kangxue Yin, Zian Wang, Wenzheng Chen, Zan Gojcic, Sanja Fidler, Nicholas Sharp, and Jun Gao.
\newblock Flexible isosurface extraction for gradient-based mesh optimization.
\newblock \emph{ACM Trans. Graph.}, 42\penalty0 (4):\penalty0 37--1, 2023.

\bibitem[Shi et~al.(2023{\natexlab{a}})Shi, Chen, Zhang, Liu, Xu, Wei, Chen, Zeng, and Su]{shi2023zero123++}
Ruoxi Shi, Hansheng Chen, Zhuoyang Zhang, Minghua Liu, Chao Xu, Xinyue Wei, Linghao Chen, Chong Zeng, and Hao Su.
\newblock Zero123++: a single image to consistent multi-view diffusion base model.
\newblock \emph{arXiv preprint arXiv:2310.15110}, 2023{\natexlab{a}}.

\bibitem[Shi et~al.(2023{\natexlab{b}})Shi, Wang, Cao, Tang, Qi, Yang, Huang, Liu, Zhang, and Shum]{shi2023toss}
Yukai Shi, Jianan Wang, He Cao, Boshi Tang, Xianbiao Qi, Tianyu Yang, Yukun Huang, Shilong Liu, Lei Zhang, and Heung-Yeung Shum.
\newblock Toss: High-quality text-guided novel view synthesis from a single image.
\newblock \emph{arXiv preprint arXiv:2310.10644}, 2023{\natexlab{b}}.

\bibitem[Shi et~al.(2024)Shi, Wang, Ye, Long, Li, and Yang]{shi2023mvdream}
Yichun Shi, Peng Wang, Jianglong Ye, Mai Long, Kejie Li, and Xiao Yang.
\newblock Mvdream: Multi-view diffusion for 3d generation.
\newblock \emph{ICLR}, 2024.

\bibitem[Shue et~al.(2023)Shue, Chan, Po, Ankner, Wu, and Wetzstein]{shue20233d}
J~Ryan Shue, Eric~Ryan Chan, Ryan Po, Zachary Ankner, Jiajun Wu, and Gordon Wetzstein.
\newblock 3d neural field generation using triplane diffusion.
\newblock In \emph{CVPR}, pages 20875--20886, 2023.

\bibitem[Song et~al.(2021{\natexlab{a}})Song, Meng, and Ermon]{song2020denoising}
Jiaming Song, Chenlin Meng, and Stefano Ermon.
\newblock Denoising diffusion implicit models.
\newblock \emph{ICLR}, 2021{\natexlab{a}}.

\bibitem[Song et~al.(2021{\natexlab{b}})Song, Sohl-Dickstein, Kingma, Kumar, Ermon, and Poole]{song2020score}
Yang Song, Jascha Sohl-Dickstein, Diederik~P Kingma, Abhishek Kumar, Stefano Ermon, and Ben Poole.
\newblock Score-based generative modeling through stochastic differential equations.
\newblock \emph{ICLR}, 2021{\natexlab{b}}.

\bibitem[Song et~al.(2023)Song, Dhariwal, Chen, and Sutskever]{song2023consistency}
Yang Song, Prafulla Dhariwal, Mark Chen, and Ilya Sutskever.
\newblock Consistency models.
\newblock \emph{ICML}, 2023.

\bibitem[Sun et~al.(2023)Sun, Zhang, Shao, Wang, Liu, Xie, and Liu]{sun2023dreamcraft3d}
Jingxiang Sun, Bo Zhang, Ruizhi Shao, Lizhen Wang, Wen Liu, Zhenda Xie, and Yebin Liu.
\newblock Dreamcraft3d: Hierarchical 3d generation with bootstrapped diffusion prior.
\newblock \emph{arXiv preprint arXiv:2310.16818}, 2023.

\bibitem[Szymanowicz et~al.(2023)Szymanowicz, Rupprecht, and Vedaldi]{szymanowicz2023splatter}
Stanislaw Szymanowicz, Christian Rupprecht, and Andrea Vedaldi.
\newblock Splatter image: Ultra-fast single-view 3d reconstruction.
\newblock \emph{arXiv preprint arXiv:2312.13150}, 2023.

\bibitem[Tang et~al.(2024{\natexlab{a}})Tang, Chen, Chen, Wang, Zeng, and Liu]{tang2024lgm}
Jiaxiang Tang, Zhaoxi Chen, Xiaokang Chen, Tengfei Wang, Gang Zeng, and Ziwei Liu.
\newblock Lgm: Large multi-view gaussian model for high-resolution 3d content creation.
\newblock \emph{arXiv preprint arXiv:2402.05054}, 2024{\natexlab{a}}.

\bibitem[Tang et~al.(2024{\natexlab{b}})Tang, Ren, Zhou, Liu, and Zeng]{tang2023dreamgaussian}
Jiaxiang Tang, Jiawei Ren, Hang Zhou, Ziwei Liu, and Gang Zeng.
\newblock Dreamgaussian: Generative gaussian splatting for efficient 3d content creation.
\newblock \emph{ICLR}, 2024{\natexlab{b}}.

\bibitem[Tang et~al.(2024{\natexlab{c}})Tang, Chen, Wang, Tang, Zhang, Fan, Chandra, Furukawa, and Ranjan]{tang2024mvdiffusion++}
Shitao Tang, Jiacheng Chen, Dilin Wang, Chengzhou Tang, Fuyang Zhang, Yuchen Fan, Vikas Chandra, Yasutaka Furukawa, and Rakesh Ranjan.
\newblock Mvdiffusion++: A dense high-resolution multi-view diffusion model for single or sparse-view 3d object reconstruction.
\newblock \emph{arXiv preprint arXiv:2402.12712}, 2024{\natexlab{c}}.

\bibitem[Tochilkin et~al.(2024)Tochilkin, Pankratz, Liu, Huang, Letts, Li, Liang, Laforte, Jampani, and Cao]{tochilkin2024triposr}
Dmitry Tochilkin, David Pankratz, Zexiang Liu, Zixuan Huang, Adam Letts, Yangguang Li, Ding Liang, Christian Laforte, Varun Jampani, and Yan-Pei Cao.
\newblock Triposr: Fast 3d object reconstruction from a single image.
\newblock \emph{arXiv preprint arXiv:2403.02151}, 2024.

\bibitem[Wang et~al.(2023)Wang, Du, Li, Yeh, and Shakhnarovich]{wang2023score}
Haochen Wang, Xiaodan Du, Jiahao Li, Raymond~A Yeh, and Greg Shakhnarovich.
\newblock Score jacobian chaining: Lifting pretrained 2d diffusion models for 3d generation.
\newblock In \emph{CVPR}, pages 12619--12629, 2023.

\bibitem[Wang et~al.(2004)Wang, Bovik, Sheikh, and Simoncelli]{wang2004image}
Zhou Wang, Alan~C Bovik, Hamid~R Sheikh, and Eero~P Simoncelli.
\newblock Image quality assessment: from error visibility to structural similarity.
\newblock \emph{IEEE TIP}, 13\penalty0 (4):\penalty0 600--612, 2004.

\bibitem[Wang et~al.(2024{\natexlab{a}})Wang, Lu, Wang, Bao, Li, Su, and Zhu]{wang2023prolificdreamer}
Zhengyi Wang, Cheng Lu, Yikai Wang, Fan Bao, Chongxuan Li, Hang Su, and Jun Zhu.
\newblock Prolificdreamer: High-fidelity and diverse text-to-3d generation with variational score distillation.
\newblock \emph{NeurIPS}, 36, 2024{\natexlab{a}}.

\bibitem[Wang et~al.(2024{\natexlab{b}})Wang, Wang, Chen, Xiang, Chen, Yu, Li, Su, and Zhu]{wang2024crm}
Zhengyi Wang, Yikai Wang, Yifei Chen, Chendong Xiang, Shuo Chen, Dajiang Yu, Chongxuan Li, Hang Su, and Jun Zhu.
\newblock Crm: Single image to 3d textured mesh with convolutional reconstruction model.
\newblock \emph{arXiv preprint arXiv:2403.05034}, 2024{\natexlab{b}}.

\bibitem[Wu et~al.(2023)Wu, Li, Yang, Zhang, Pan, Wang, Lin, and Liu]{wu2023hyperdreamer}
Tong Wu, Zhibing Li, Shuai Yang, Pan Zhang, Xingang Pan, Jiaqi Wang, Dahua Lin, and Ziwei Liu.
\newblock Hyperdreamer: Hyper-realistic 3d content generation and editing from a single image.
\newblock In \emph{SIGGRAPH Asia 2023 Conference Papers}, pages 1--10, 2023.

\bibitem[Xie et~al.(2023)Xie, Li, Tan, Sun, Shu, Zhou, Bi, Pirk, and Kaufman]{xie2023carve3d}
Desai Xie, Jiahao Li, Hao Tan, Xin Sun, Zhixin Shu, Yi Zhou, Sai Bi, S{\"o}ren Pirk, and Arie~E Kaufman.
\newblock Carve3d: Improving multi-view reconstruction consistency for diffusion models with rl finetuning.
\newblock \emph{arXiv preprint arXiv:2312.13980}, 2023.

\bibitem[Xu et~al.(2024{\natexlab{a}})Xu, Cheng, Gao, Wang, Gao, and Shan]{xu2024instantmesh}
Jiale Xu, Weihao Cheng, Yiming Gao, Xintao Wang, Shenghua Gao, and Ying Shan.
\newblock Instantmesh: Efficient 3d mesh generation from a single image with sparse-view large reconstruction models.
\newblock \emph{arXiv preprint arXiv:2404.07191}, 2024{\natexlab{a}}.

\bibitem[Xu et~al.(2024{\natexlab{b}})Xu, Tan, Luan, Bi, Wang, Li, Shi, Sunkavalli, Wetzstein, Xu, et~al.]{xu2023dmv3d}
Yinghao Xu, Hao Tan, Fujun Luan, Sai Bi, Peng Wang, Jiahao Li, Zifan Shi, Kalyan Sunkavalli, Gordon Wetzstein, Zexiang Xu, et~al.
\newblock Dmv3d: Denoising multi-view diffusion using 3d large reconstruction model.
\newblock \emph{ICLR}, 2024{\natexlab{b}}.

\bibitem[Yeh et~al.(2024)Yeh, Huang, Kim, Xiao, Nguyen-Phuoc, Khan, Zhang, Chandraker, Marshall, Dong, et~al.]{yeh2024texturedreamer}
Yu-Ying Yeh, Jia-Bin Huang, Changil Kim, Lei Xiao, Thu Nguyen-Phuoc, Numair Khan, Cheng Zhang, Manmohan Chandraker, Carl~S Marshall, Zhao Dong, et~al.
\newblock Texturedreamer: Image-guided texture synthesis through geometry-aware diffusion.
\newblock \emph{arXiv preprint arXiv:2401.09416}, 2024.

\bibitem[Yin et~al.(2023)Yin, Gharbi, Zhang, Shechtman, Durand, Freeman, and Park]{yin2023one}
Tianwei Yin, Micha{\"e}l Gharbi, Richard Zhang, Eli Shechtman, Fredo Durand, William~T Freeman, and Taesung Park.
\newblock One-step diffusion with distribution matching distillation.
\newblock \emph{arXiv preprint arXiv:2311.18828}, 2023.

\bibitem[Yu et~al.(2021)Yu, Ye, Tancik, and Kanazawa]{yu2021pixelnerf}
Alex Yu, Vickie Ye, Matthew Tancik, and Angjoo Kanazawa.
\newblock pixelnerf: Neural radiance fields from one or few images.
\newblock In \emph{CVPR}, pages 4578--4587, 2021.

\bibitem[Yu et~al.(2023)Yu, Xu, Zhang, Liu, Ye, Wu, Yan, Zhu, Xiong, Liang, et~al.]{yu2023mvimgnet}
Xianggang Yu, Mutian Xu, Yidan Zhang, Haolin Liu, Chongjie Ye, Yushuang Wu, Zizheng Yan, Chenming Zhu, Zhangyang Xiong, Tianyou Liang, et~al.
\newblock Mvimgnet: A large-scale dataset of multi-view images.
\newblock In \emph{CVPR}, pages 9150--9161, 2023.

\bibitem[Zhang(2022)]{zhang2022reference}
Lyuming Zhang.
\newblock Reference-only control, 2022.

\bibitem[Zhang et~al.(2023)Zhang, Rao, and Agrawala]{zhang2023controlnet}
Lvmin Zhang, Anyi Rao, and Maneesh Agrawala.
\newblock Adding conditional control to text-to-image diffusion models.
\newblock In \emph{ICCV}, 2023.

\bibitem[Zhang et~al.(2018)Zhang, Isola, Efros, Shechtman, and Wang]{zhang2018unreasonable}
Richard Zhang, Phillip Isola, Alexei~A Efros, Eli Shechtman, and Oliver Wang.
\newblock The unreasonable effectiveness of deep features as a perceptual metric.
\newblock In \emph{CVPR}, pages 586--595, 2018.

\bibitem[Zheng et~al.(2024)Zheng, Lu, Chen, and Zhu]{zheng2024dpm}
Kaiwen Zheng, Cheng Lu, Jianfei Chen, and Jun Zhu.
\newblock Dpm-solver-v3: Improved diffusion ode solver with empirical model statistics.
\newblock \emph{NeurIPS}, 36, 2024.

\bibitem[Zhu et~al.(2024)Zhu, Zhuang, and Koyejo]{zhu2023hifa}
Junzhe Zhu, Peiye Zhuang, and Sanmi Koyejo.
\newblock Hifa: High-fidelity text-to-3d generation with advanced diffusion guidance.
\newblock In \emph{ICLR}, 2024.

\bibitem[Zhuang et~al.(2023)Zhuang, Wang, Lin, Liu, and Li]{zhuang2023dreameditor}
Jingyu Zhuang, Chen Wang, Liang Lin, Lingjie Liu, and Guanbin Li.
\newblock Dreameditor: Text-driven 3d scene editing with neural fields.
\newblock In \emph{SIGGRAPH Asia 2023 Conference Papers}, pages 1--10, 2023.

\bibitem[Zou et~al.(2024)Zou, Yu, Guo, Li, Liang, Cao, and Zhang]{zou2023triplane}
Zi-Xin Zou, Zhipeng Yu, Yuan-Chen Guo, Yangguang Li, Ding Liang, Yan-Pei Cao, and Song-Hai Zhang.
\newblock Triplane meets gaussian splatting: Fast and generalizable single-view 3d reconstruction with transformers.
\newblock \emph{CVPR}, 2024.

\end{thebibliography}
}

\appendix

\section{Datasets}
\topic{Objaverse}
We use the Objaverse 1.0~\cite{deitke2023objaverse} LVIS subset\footnote{\url{https://objaverse.allenai.org/docs/objaverse-1.0\#lvis-annotations}} for our training, which contains around 46K 3D objects. Since our distillation method only requires the condition image as the input, we render one image for each scene at a random viewpoint with $49.1$ FOV and a camera radius of $1.5$.

\topic{GSO} We use the same randomly selected 30 objects as in ~\cite{liu2023syncdreamer} from Google Scanned Objects (GSO)~\cite{downs2022google} dataset for evaluation. For each object, we use blender scripts to render an image with a size of $512 \times 512$ as the input view with zero elevation and azimuth.
We render another two sets of images for evaluation: the first consists of 6 images from the same viewpoint as in Zero123Plus~\cite{shi2023zero123++}, and the second consists of evenly sampled 15 images around the object with zero elevation.

\section{Implementation Details}
\subsection{Pretrained Models}
\topic{Zero123Plus Model}
We use the white background Zero123Plus v1.1 model fine-tuned by InstantMesh~\cite{xu2024instantmesh} without the depth ControlNet~\cite{zhang2023controlnet} part. The input to the model is a white background image and the output is a set of six novel view images at elevation $(30, -20, 30, -20, 30, -20)$ and azimuth $(30, 90, 150, 210, 270, 330)$. The elevation angles are absolute and the azimuth angles are relative to the input view. During inference, Zero123Plus forwards the UNet two times. For the first time, the UNet takes the condition image as input and the keys and values matrices of the self-attention matrices are stored. For the second time, the UNet takes the noisy image as input and concatenates the stored self-attention matrices for conditioning. 

\topic{InstantMesh} 3D meshes~\cite{kerbl20233dgs} are reconstructed from six input images at random poses of an input object. It mainly adopts the transformer architecture in LRM~\cite{openlrm, hong2023lrm} with two major differences: first, the input of the network is extended from single-view to six-views; second, the output 3D representation is changed from triplanes to a differentiable iso-surface extraction module, \ie, Flexicubes~\cite{shen2023flexible} to enable efficient rendering and mesh extraction. 

\subsection{Training Details}
\sysname{} can be trained quite efficiently, here we provide the details our training:

\topic{Stage I: Multi-view Score Distillation} 
When we train the multi-view generator, we have three UNets that all initialized with the same architecture and parameters. The teacher model is freezing, and the multi-view generator and the student model are trained.
The training procedure is similar to the standard VSD training implemented in threestudio~\cite{liuthreestudio}.
For each iteration, we first random sample a batch of Gaussian noises $\bz$ and use it as the multi-view generator input. The generator output is sent to the pretrained teacher and student model for VSD loss. The gradients are backpropagated to the generator. Afterward, we add noises at level $t$ in $[0.02, 0.98]$ to the generator output and train the student model to predict the added noises. We follow the original Zero123Plus~\cite{shi2023zero123++} model and use the DDPM Scheduler~\cite{ho2020denoising} with $v$-prediction to train the student model. The training takes about 4 hours on a single NVIDIA L40 GPU.

\topic{Stage II: 3D Consistent Distillation}
For this part, we follow the InstantMesh~\cite{xu2024instantmesh} architecture design.
During training, we use a batch size of 4 on 4 NVIDIA L40 GPUs for 5 epochs. The whole training takes about 10 hours. For each scene, we always use the fixed 6 viewpoints from the multi-view diffusion model as input and randomly sample another 4 views from the pseudo ground truth set to compute loss.

\section{Baselines}
We compare with other methods that can achieve 3D generation in a feed-forward manner. Current works that can achieve this goal are reconstruction-based methods, which train a model to predict novel views given an input image with a regression loss. We select the most representative and state-of-the-art methods for comparison: LRM~\cite{hong2023lrm}, TriplaneGaussian~\cite{zou2023triplane} and TripoSR~\cite{tochilkin2024triposr}.

\topic{LRM} trains a large-scale transformer model that uses the attention operations between learnable input embeddings and input image features to directly output triplanes.
Since the official code is not public, we use the community open-source version OpenLRM~\cite{openlrm} for comparison\footnote{\url{https://github.com/3DTopia/OpenLRM}}. We use the openlrm-mix-large-1.1 model trained on Objaverse~\cite{deitke2023objaverse} and MVImgNet~\cite{yu2023mvimgnet} datasets as the comparison.

\topic{TriplaneGaussian} utilizes two transformer-based networks: a point decoder and a triplane decoder, to reconstruct 3D objects. The triplane features and point features are combined to decode 3D Gaussians for fast novel view synthesis.
We use the official code for evaluation\footnote{\url{https://github.com/VAST-AI-Research/TriplaneGaussian}}. 

\topic{TripoSR} follows the design principle of LRM and uses a transformer network to predict triplanes from a single image. The main difference is that TripoSR curated and rendered a new set of 3D object data and employed mask loss and patched rendering loss. We use the official code for evaluation\footnote{\url{https://github.com/VAST-AI-Research/TripoSR}}.

\begin{figure*}[htbp]
    \centering
    \includegraphics[width=1.\linewidth]{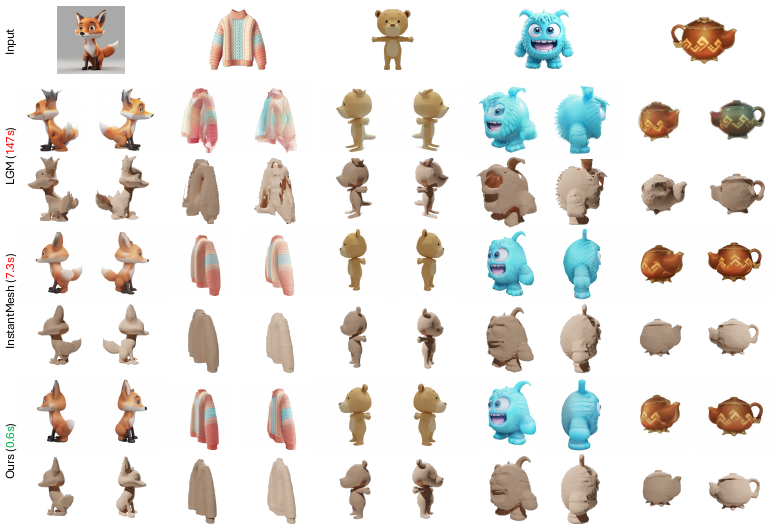}
    \caption{Qualitative comparison of our method and LGM, InstantMesh, the two columns for each method are the RGB and geometry renderings respectively.}
    \label{fig:suppcomp}
\end{figure*}

\begin{figure*}[tp]
    \centering
    \includegraphics[width=1.\linewidth]{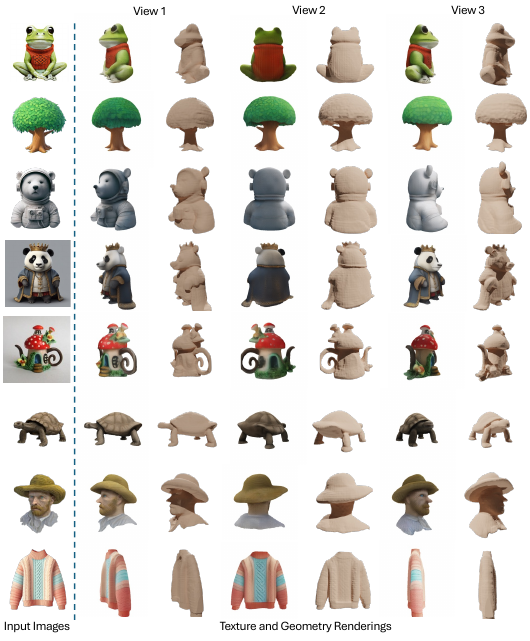}
    \caption{Additional results of our method.}
    \label{fig:suppres}
\end{figure*}

\section{Additional Results}
\subsection{Qualitative Comparison with LGM and InstantMesh}
Besides comparing with feedforward baselines, we also compare with methods that use multi-view diffusion models and multi-view reconstruction models for 3D generation: LGM~\cite{tang2024lgm} and InstantMesh~\cite{xu2024instantmesh}. LGM first predicts multi-view images and then uses an asymmetric UNet architecture that outputs 3D Gaussians stored by splatter images~\cite{szymanowicz2023splatter} of size $128 \times 128 \times 14$ from $256 \times 256$ input images. The $14$ channels of splatter images all the parameters of 3D Gaussians, including color, position, rotation, scale, \etc. The whole process takes less than 2 seconds to generate 3D Gaussians, but it takes about 2 minutes to extract meshes from the Gaussians. InstantMesh is our teacher model that uses 75 steps of Zero123Plus to generate multi-view images and produces meshes directly from these images.
The qualitative results shown in \cref{fig:suppcomp} show that LGM produces multi-view inconsistent renderings, floating artifacts and bad geometry, presumably due to the 3D Gaussian representation. InstantMesh generates sharp renderings and smooth surfaces. Our method also has comparable results with our teacher InstantMesh, while being more than 10 times faster.

\subsection{2D Renderings}
More renderings of the 3D meshes generated by our method can be found in \cref{fig:suppres}.




\end{document}